%% file: acl_latex.tex
\pdfoutput=1
\documentclass[11pt]{article}

\usepackage{acl}

\usepackage{times}
\usepackage{latexsym}

\usepackage[T1]{fontenc}
\usepackage[utf8]{inputenc}

\usepackage{microtype}

\usepackage{amsmath}
\usepackage{mathtools}
\usepackage[ruled,vlined,noend]{algorithm2e}
\usepackage{graphicx}
\usepackage{mathtools}
\usepackage{bbm}
\usepackage{amsfonts}
\usepackage[inline]{enumitem}
\usepackage{tabularx}
\usepackage{arydshln}
\usepackage{multirow}
\usepackage{makecell}
\usepackage{xcolor}
\usepackage{soul}
\usepackage[off]{svg-extract}
\svgsetup{clean=true}
\usepackage[inline]{enumitem}
\usepackage{adjustbox}

\SetKwIF{If}{ElseIf}{Else}{if}{}{else if}{else}{end if}%
\SetKwFor{While}{while}{}{end while}%
\SetKwRepeat{Do}{do}{while}
\SetKw{KwGoTo}{go to}
\usepackage{mathtools}

\newif\ifcomments

\providecommand{\commentout}[1]{}

\newcommand\HPT{HPT}
\newcommand\hpt{hyperparameter tuning}
\newcommand\gof{goodness-of-fit}
\newcommand\RSQTH{0.95}

\newcommand\bestFone{\emph{Best-$F_1$}}



\title{Scaling Laws Under the Microscope: \\Predicting Transformer Performance from Small Scale Experiments}

\author{Maor Ivgi ~~~~~~Yair Carmon ~~~~~~
Jonathan Berant \\
The Blavatnik School of Computer Science, Tel-Aviv University \\
\small{\texttt{\{maor.ivgi,joberant\}@cs.tau.ac.il~,~ycarmon@tauex.tau.ac.il}}} 

\begin{document}
\maketitle

\input{00_abstract}
\input{01_intro}
\input{02_method}

\input{03_upstream}
\input{04_downstream}

\input{05_related_work}
\input{06_discussion}
\input{07_conclusions}
\input{08_limitations}
\input{09_acknowledgements}

\bibliography{custom}
\bibliographystyle{acl_natbib}

\appendix
\input{appendix_00_main}

\end{document}

%% file: 00_abstract.tex
\begin{abstract}
Neural scaling laws define a predictable relationship between a model's parameter count and its performance after training in the form of a power law. However, most research to date has not explicitly investigated whether scaling laws can be used to accelerate model development. 
In this work, we perform such an empirical investigation across a wide range of language understanding tasks, starting from models with as few as 10K parameters, and evaluate downstream performance across 9 language understanding tasks.
We find that scaling laws emerge at finetuning time in \emph{some} NLP tasks, and that they can also be exploited for debugging convergence when training large models. Moreover, for tasks where scaling laws exist, they can be used to predict the performance of larger models, which enables effective model selection. However, revealing scaling laws
requires careful hyperparameter tuning and multiple runs for the purpose of uncertainty estimation, which incurs additional overhead, partially offsetting the computational benefits.

\end{abstract}

%% file: 01_intro.tex
\section{Introduction}

\input{figure_defs/figure_intro}
Transformer-based language models (LMs) \cite{Vaswani2017AttentionIA, Devlin2019BERTPO,Raffel2020ExploringTL,Radford2019LanguageMA,Brown2020LanguageMA} which are at the foundation of modern NLP systems, have been recently shown to exhibit scaling laws \cite{Kaplan2020ScalingLF, Hernandez2021ScalingLF,Tay2021ScaleEI,Tay2022ScalingLV}, that is, the test loss of LMs obeys a predictable power law with respect to the number of model parameters, dataset size, and computation budget. This finding ignited substantial research that demonstrated scaling laws in a wide range of areas including computer vision \cite{Rosenfeld2020ACP,Henighan2020ScalingLF,Zhai2021ScalingVT,Bahri2021ExplainingNS,abnar2021exploring}, acoustic models \cite{Droppo2021ScalingLF}, and board games \cite{Jones2021ScalingSL,BenAssayag2021TrainOS}, among others. 

On top of being a fascinating phenomenon on its own, scaling laws can potentially be harnessed for more efficient development of models. Specifically, if scaling laws hold, one can perform modeling decisions at multiple small scales, and extrapolate to infer which model will perform best at a larger scale. 
While starting small is an established technique \cite{Tan2019EfficientNetRM}, scaling laws can provide a more principled framework for this methodology.
Moreover, scaling laws can potentially accelerate the development cycle, and reduce carbon footprint caused by training large neural models \cite{Schwartz2020GreenA,Bender2021OnTD}. 

However, for this idea to materialize, several questions must be addressed, which have not been fully considered  by past literature.
First, do scaling laws consistently occur at finetuning time across a wide range of language understanding tasks? Second, do they manifest reliably at the small-scale regime? And third, what is the predictive power of scaling laws when used to predict the behavior of large models for the purpose of model selection? While recent work touched upon these questions \cite{Rosenfeld2020ACP, Tay2021ScaleEI,Tay2022ScalingLV}, most work focused on post-hoc analysis of highly-parameterized models, 
and looked at performance on aggregate benchmarks, such as GLUE \cite{Wang2018GLUEAM} and SuperGLUE \cite{Wang2019SuperGLUEAS}, without analyzing performance at the level of individual dataset.

In this work, we perform a thorough empirical evaluation of scaling laws from the perspective of NLP researchers with resource constraints, addressing the aforementioned questions. We analyze scaling laws at finetuning time across 9 different tasks with models ranging from merely 10K parameters up to roughly 100M (e.g. BERT-Base \cite{Devlin2019BERTPO}).
Moreover, we move away from post-hoc analysis and evaluate whether scaling laws can be used to predict the performance of larger models. As a case study, we test whether scaling laws can be used for model selection, by  comparing the performance of two LMs with two different pretraining objectives: Masked Language Modeling (MLM) \cite{Devlin2019BERTPO} and Pointwise Mutual Information (PMI) \cite{levine2020pmi}.

Our experiments reveal several findings. 
\begin{enumerate}
\item We show near-perfect scaling laws with high \gof\ at pretraining time for both MLM and PMI even with the smallest models and for multiple architectural choices (\S\ref{sec:pretraining}). 

\item At finetuning time, scaling laws emerge only in some of the tasks (\S\ref{subsec:born-equal}).
Fig.~\ref{fig:intro} (top) shows an example of a good fit over SQuAD \cite{rajpurkar2016squad} and a less impressive fit over MRPC \cite{dolan2005automatically}.

\item In some tasks, a certain minimal model size is required (\S\ref{subsec:born-equal}). For example, on MRPC only models with at least four hidden layers performed significantly better than chance.

\item Careful \hpt{} (\HPT) can be crucial for exposing precise scaling laws at finetuning time, especially at smaller scales (\S\ref{subsec:hpt}). For example, in Fig.~\ref{fig:intro} (bottom), not only is the fit much better with \HPT\ (blue), it also dramatically affects the prediction at larger scales. 

\item As for using scaling laws for model selection (\S\ref{subsec:predictive},\S\ref{subsec:model-selection}), our MLM vs. PMI case study shows that scaling laws can be used to perform model selection at larger scales whenever we observe high goodness-of-fit ($R^2>\RSQTH$) of the scaling law at smaller scales.

\end{enumerate}

Overall, our empirical findings paint a nuanced picture of the potential of scaling laws as a tool for model design. On one hand, we observe scaling laws at finetuning time for some NLP tasks, and show that they can be used to predict the performance of a model that is 10x larger. On the other hand, this does not happen consistently on all tasks, and revealing these scaling laws requires careful control over hyperparameters and convergence conditions, incurring additional overhead that might counteract the computational benefits.

%% file: figure_defs/figure_intro.tex
\begin{figure}[t]
\begin{center}

\centerline{
\includegraphics[width=\linewidth]{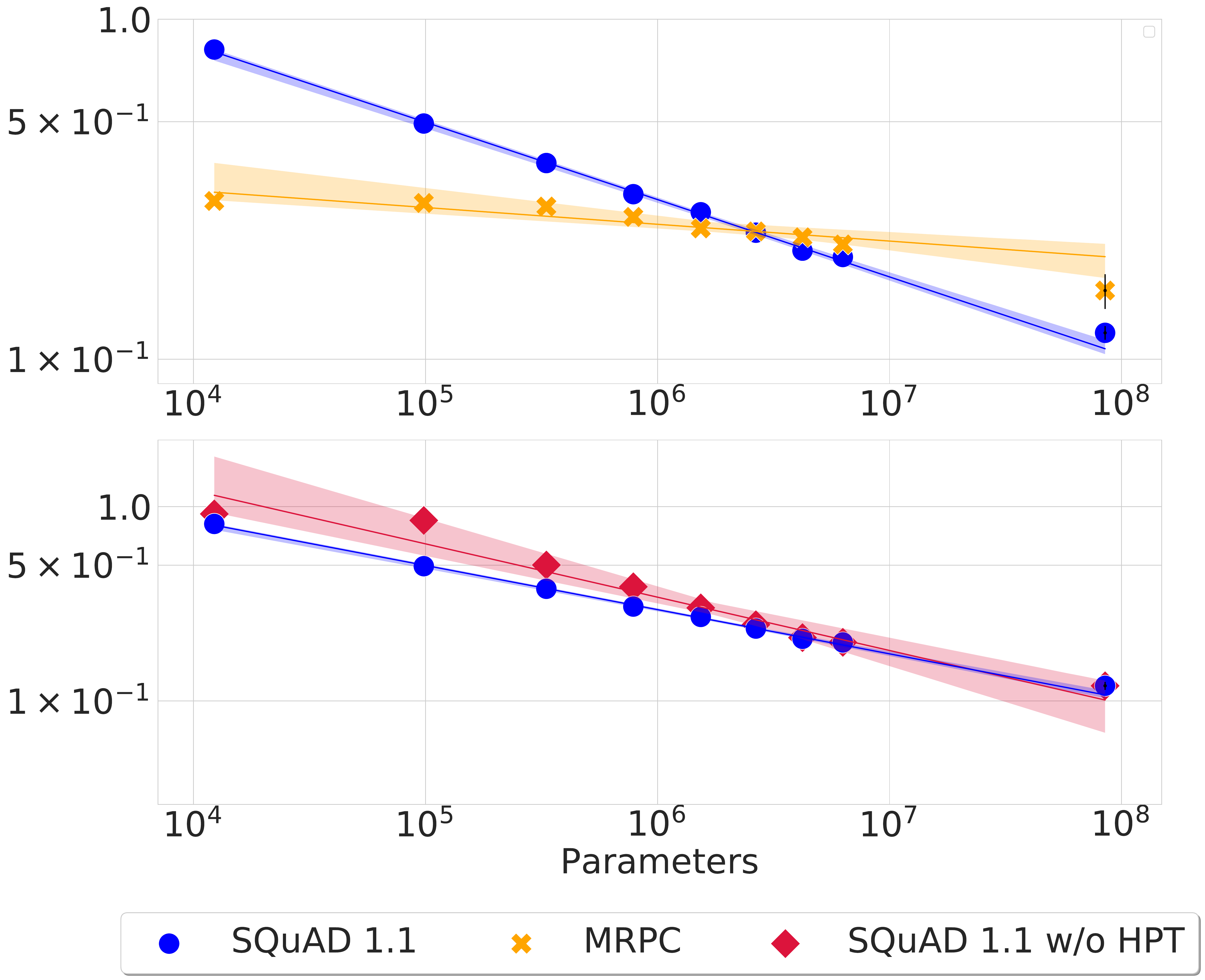}
}
\setlength{\belowcaptionskip}{-5pt}
\caption{
Performance of models when finetuned on SQuAD 1.1 (evaluated using $1-F_1$) and MRPC (classification error).
\textbf{Top}: Models exhibit a clean scaling law fit on SQuAD 1.1 ($R^2=0.998$) 
compared to MRPC ($R^2=0.763$).
\textbf{Bottom}: Without \hpt{} (w/o \HPT) the \gof{} is lower ($R^2=0.83$).
}

\label{fig:intro}
\end{center}
\end{figure}

%% file: 02_method.tex
\section{Method}
\label{sec:method}
We describe our experimental setup in §\ref{subsec:exp_setup}, and then our procedure for evaluating \gof\ and the predictive power of scaling laws in §\ref{subsec:eval}.

\subsection{Experimental setup}
\label{subsec:exp_setup}

\paragraph{Architecture} We consider encoder-only transformer models, similar to the architecture of BERT \cite{Devlin2019BERTPO} and RoBERTa \cite{liu2019roberta}. 
Encoder-only models are ubiquitous in NLP for a wide range of classification tasks such as natural language inference, question answering, information extraction, text classification, and more.

\paragraph{Model configurations} Recent work \cite{Tay2022ScalingLV, Tay2021ScaleEI, Hernandez2021ScalingLF,Zhai2021ScalingVT,Bahri2021ExplainingNS} focused on parameter-rich transformers, ranging from 5--10M trainable parameters to 40B.
We, instead, investigate smaller models, with as few as 10K trainable parameters, assuming a computationally-constrained environment.
To preserve the architecture as we scale model parameters, we increase the number of layers ($L$) from one to twelve, while keeping the aspect ratio (AR) constant ($\text{AR}:=\frac{L}{H}$ where $H$ is the hidden layer width).

We experiment with two different aspect ratios, 32 and 64, and scale our models over four orders of magnitude, up to the $\sim85M$ parameters of BERT-Base \cite{Devlin2019BERTPO}. In particular, we train small-scale models with aspect ratio 32 with 1 to 8 layers, and small-scale models with aspect ratio 64 with 1 to 5 layers. Finally, we train from scratch a BERT-Base \cite{Devlin2019BERTPO} model which has 12 hidden layers and aspect ratio 64. A detailed list of all configurations is in Table~\ref{table:model-configurations} in App.~\ref{app:experimental-details}.

\paragraph{Pretraining} 
We experiment with two pretraining objectives (i.e., ``upstream''), as different objectives can affect models downstream performance \cite{Devlin2019BERTPO,liu2019roberta,Raffel2020ExploringTL,levine2020pmi}.
As a test case, 
we compare the popular Masked Language Modeling (MLM) masking \cite{Devlin2019BERTPO}, where random tokens are masked, to Pointwise Mutual Information (PMI) masking \cite{levine2020pmi}, where masking is performed over sequences of tokens that tend to co-occur in the training corpus.
We test whether small scale experiments on these objectives can predict performance at larger scale and inform which objective is better for a particular task.
To the best of our knowledge, prior work did not compare the effects of different pretraining objectives on the behavior of scaling laws.

\input{table_defs/datasets}

\paragraph{Finetuning tasks} Past work \cite{Henighan2020ScalingLF,abnar2021exploring,Zhai2021ScalingVT,Bahri2021ExplainingNS} evaluated performance on finetuning (i.e., ``downstream'') tasks in computer vision, but less attention has been given to the relation between architectures and finetuning accuracy in NLP.
In this work, we address this lacuna and report results after finetuning on 9 different datasets:
SQuAD 1.1 \cite{rajpurkar2016squad}, MNLI \cite{Williams2018ABC}, QNLI \cite{Wang2018GLUEAM}, SST-2 \cite{socher2013recursive}, RACE \cite{lai2017race}, CoLA \cite{warstadt2018neural}, SQuAD 2.0 \cite{rajpurkar2018know}, MRPC \cite{dolan2005automatically} and BoolQ \cite{clark2019boolq}.

For all tasks, we use the common classification head (single layer MLP) on top of the prepended \texttt{CLS} token. In each task, we evaluate using the official metric and finetune on the training data suggested by the authors, except in MNLI where we randomly sample a subset of 25\% of the training data for efficiency.
The full specification of datasets can be found in Table~\ref{table:datasets}. For details on the finetuning procedure, refer to App.~\ref{app:experimental-details}. 

\paragraph{Notation} 
\label{subsec:notation}
Similar to \newcite{Kaplan2020ScalingLF}, we denote the number of trainable parameters, not including word embeddings, by $N$, and estimate it with $N \approx 12LH^2$ where $L$ is the number of layers and $H$ is the hidden dimension.

\subsection{Evaluation}
\label{subsec:eval}

While past work reported the scaling coefficients of a fitted power law, there is no consensus on a measure for estimating the \emph{\gof} of a scaling law to a set of points. Moreover, once the power law is computed, less attention has been dedicated to estimating its \emph{predictive power} to larger scales. We propose evaluation metrics for these quantities.

\paragraph{Goodness-of-fit} Given a task and a metric $F: \mathcal{X} \rightarrow \mathbb{R}_{\geq 0}$ to be minimized, we analyze an architecture across $M$ different scales, where we finetune the architecture $T$ times per scale.
To find the power law coefficients, we fit a regression line by performing \emph{least-squares} in log-log scale over all $M\cdot T$ points. 
We then define the \gof\ as the $R^2$ measure given by the line and the points.

Because different seeds result in different performance, we wish to compute confidence intervals for the regression line. To this end, we employ a hierarchical bootstrap procedure, where we sample a set of data points $B = 1000$ times, use the points to produce $B$ fitted lines, and compute confidence intervals ($[2.5,97.5]$ percentiles) around the slope and around each point along the line. We provide full details on the hierarchical bootstrap procedure and the computation of $R^2$ in App.~\ref{app:evaluation}.

\paragraph{Predictive power} Given data on the performance of a model on a larger scale, we evaluate the predictive power of the inferred scaling law by computing the \emph{Mean-Relative-Error (MRE)}
between the predicted performance and the true one. In particular, given the true performance values of $k$ experiments $y=(y_1,\dots,y_k)$ on $k$ larger scales, and the corresponding predictions $\hat{y}=(\hat{y}_1,\dots,\hat{y}_k)$ we define 
\begin{equation}
\textit{MRE}(y,\hat{y}):=\frac{1}{k}\sum_{i=1}^{k}{\left | \frac{y_i-\hat{y}_i}{y_i}\right |}
\end{equation}
When $k=1$, we drop the absolute value to keep information on whether the model is overshooting or undershooting and call this \emph{Relative Error} (RE).

%% file: table_defs/datasets.tex
\begin{table}[t]
\centering
\footnotesize

\begin{tabular}{@{\extracolsep{0.1pt}}lcccc} 

\Xhline{2\arrayrulewidth}

  & \boldmath{$|\textbf{Train}|$} & \boldmath{$|\textbf{Eval}|$} & \textbf{Metric} & \textbf{Top freq.} \\ \cline{2-2}\cline{3-3}\cline{4-4}\cline{5-5}

\rule{-2pt}{2.6ex}

\input{figures/datasets.txt}

\Xhline{2\arrayrulewidth}

\end{tabular}
\setlength{\belowcaptionskip}{-10pt}
\caption{Statistics on downstream tasks.
\textbf{Top freq.} shows the performance when always predicting the most frequent class in the evaluation set, where applicable. See the original papers definitions of the evaluation metrics. In MNLI we use 25\% of the training examples, and use the validation-matched set as evaluation data.
}
\label{table:datasets}
\end{table}

%% file: figures/datasets.txt
 \textbf{SQuAD 1.1} &  87,599 & 10,570 &      $F_1$ &  N/A \\
      \textbf{MNLI} &  98,175 &  9,815 &       Acc. & 0.35 \\
      \textbf{QNLI} & 104,743 &  5,463 &       Acc. & 0.51 \\
     \textbf{SST-2} &  67,349 &    872 &       Acc. & 0.51 \\
      \textbf{RACE} &  87,866 &  4,887 &       Acc. & 0.27 \\
      \textbf{CoLA} &   8,551 &  1,043 &        MCC &    0 \\
 \textbf{SQuAD 2.0} & 130,319 & 11,873 &  \bestFone &  N/A \\
      \textbf{MRPC} &   3,668 &    408 &       Acc. & 0.68 \\
     \textbf{BoolQ} &   9,427 &  3,270 &       Acc. & 0.62 \\

%% file: 03_upstream.tex
\input{figure_defs/pretraining_results}

\section{Pretraining}
\label{sec:pretraining}

We pretrain all models from scratch on the \textit{Datasets} \cite{lhoest-etal-2021-datasets} Wikipedia dump, see experimental details in App.~\ref{app:experimental-details}. We use early stopping to declare convergence, but note that determining convergence is non-trivial, as we discuss in \S\ref{subsec:debugging}.

Fig.~\ref{fig:pretraining-results} shows the evaluation loss of each configuration as a function of the parameter count for both MLM and PMI. As can be seen from the clean linear relationships (in log-log scale), both aspect ratios and both objectives present a power law, with $R^2$ exceeding 0.99 in all cases. This is consistent with the language modeling results of \newcite{Kaplan2020ScalingLF}, and shows that scaling laws exist at pretraining time for the MLM and PMI objectives.

Past work \cite{Kaplan2020ScalingLF, Tay2022ScalingLV} 
examined scaling laws of different architectures (e.g., Transformer \cite{Vaswani2017AttentionIA}, Reformer \cite{Kitaev2020ReformerTE}, Performer \cite{Zhao2019PerformerSI}). They showed that while different architectures affect scaling laws,  architectural hyperparameters, such as aspect ratio (AR), make little difference as long as they are within a reasonable range.
Interestingly, in both MLM and PMI the slope of AR 64 is slightly better than the slope of AR 32, even when taking into account the slope's 95\% confidence intervals. 
The intersection of the two AR lines illustrates the potential of using scaling laws for model selection:
choosing the AR based on small scale experiments would lead to choosing AR 32, while the performance of models with AR 64 seems comparable and perhaps better in the larger scale.
However, the confidence intervals of the fit, depicted as sleeves in Fig.~\ref{fig:pretraining-results}, intersect at the larger scales, meaning we cannot predict a performance difference with confidence.

We note that our largest MLM model, which uses AR 64, 12 layers, and has 85M trainable parameters performs slightly worse than predicted by the power law, which might hint that it is under-trained.
We leave verifying this and training larger models with different ARs to future work.

\subsection{Debugging convergence with scaling laws} 
\label{subsec:debugging} 

A useful side-effect of the clean scaling law behavior during pretraining is the ability to detect issues in pretraining convergence. In several cases, training stopped due to early stopping, but its loss was greater than predicted by the fit done on other scales. When investigated further, we found that increasing the patience hyperparameter led to further significant loss reduction on the evaluation data, finally converging at the predicted value.

This result points to a methodological issue in current literature, where researchers train models ``until convergence''.
Convergence is not well-defined, since it is affected by early stopping hyperparameters, such as \emph{patience} and \emph{minimal decrease}. This can lead to under-optimized models, as we observed here. We believe that being precise about the definition of a ``converged model'' is important for reproducibility of scaling laws research. We further illustrate this and provide the precise criteria we used to declare convergence in App.~\ref{app:convergence}. 

%% file: figure_defs/pretraining_results.tex
\begin{figure*}[t]
\begin{center}
\includegraphics[width=\textwidth,trim={0 0 0 0},clip]{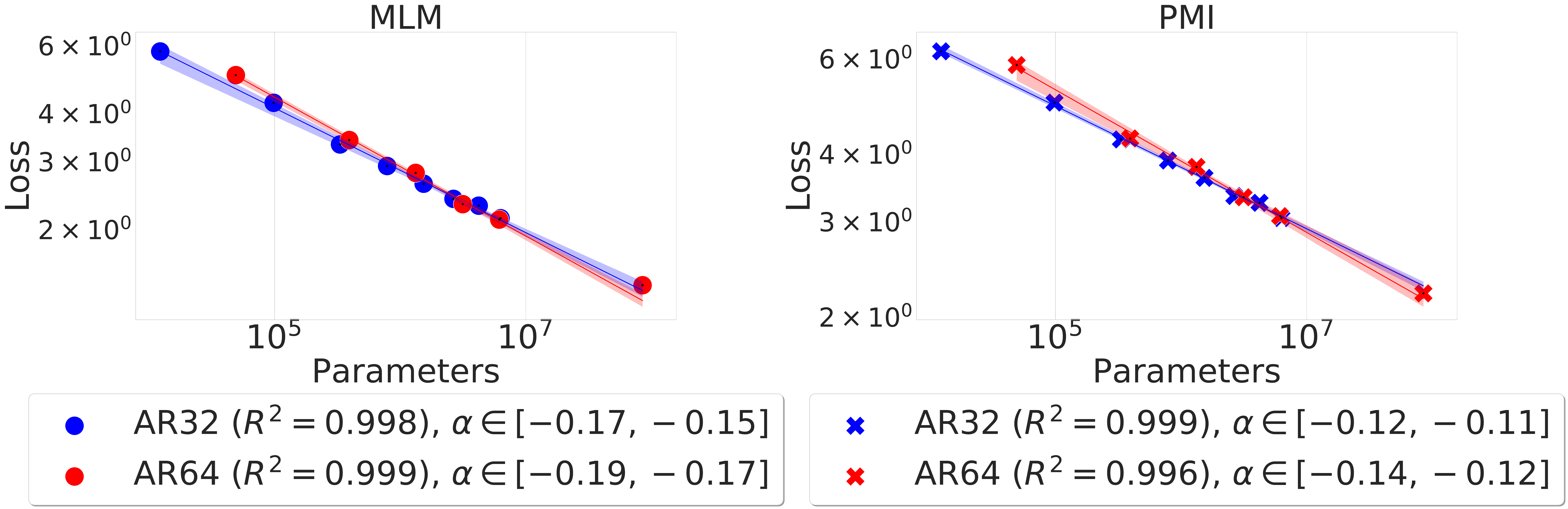}
\setlength{\belowcaptionskip}{0pt}
\setlength{\abovecaptionskip}{-5pt}
\caption{
Evaluation loss w.r.t.\ models' parameter count, colored by aspect ratio, along with \gof{} and the slope ($\alpha$). \textbf{Left}: MLM objective. \textbf{Right}: PMI objective. Note: we do not consider the BERT-Base model (rightmost point) when computing $R^2$, since we later use it to examine predictive power. Here, $M=8$ for $\text{AR}=32$ and $M=5$ for $\text{AR}=64$, and $T=1$ in both cases (see text for definitions of $M$ and $T$).
}
\label{fig:pretraining-results}
\end{center}
\end{figure*}

%% file: 04_downstream.tex
\input{figure_defs/finetuning_results}

\section{Finetuning}
\label{sec:finetuning}

\newcite{Tay2022ScalingLV} recently showed that evaluation loss during pretraining does not necessarily correlate with performance on downstream tasks.
In this section, we revisit this finding. In particular. we investigate: a) differences in scaling laws across tasks, b) the effects of \hpt, architectural design and pretraining objectives, and c) the predictive power of emerging scaling laws. All finetuning experiments use the final checkpoints of the pretrained models described in \S\ref{sec:pretraining}.

\subsection{Downstream tasks are not born equal}
\label{subsec:born-equal}
\newcite{Tay2021ScaleEI} and \newcite{Tay2022ScalingLV} showed the behavior of transformers over the aggregated GLUE \cite{Wang2018GLUEAM} and SuperGLUE \cite{Wang2019SuperGLUEAS} benchmarks. 
We now examine their behavior over a diverse set of NLP tasks. 

\input{table_defs/finetuning_results}

Fig.~\ref{fig:finetuning-results} and the $R^2$ values in Table~\ref{table:finetuning-results} show that different tasks vary in terms of the quality of the power law fit. As before, We do not consider the BERT-Base model with 85M parameters when computing $R^2$ since we use this model to test the predictive power of scaling laws.
In some tasks, such as SQuAD 1.1, MNLI and QNLI , we observe a relatively clean power law ($R^2 > 0.96$), even though their evaluation metrics (e.g., $F_1$ and accuracy) differ from their finetuning loss. 
On the other end of the spectrum is BoolQ, which is not even monotonic w.r.t.\ number of parameters.
Other tasks lie in different places along this spectrum.

We hypothesize that the two factors that play a role in determining the emergence of scaling laws during finetuning are (a) the proximity of the task to the pretraining objective, and (b) the amount of data to finetune on (Table~\ref{table:datasets}).
Namely, on all tasks where the training data contains less than 10K examples, $R^2$ was low ($<0.85$). Furthermore, 
in tasks where $R^2 > \RSQTH$, the objective can be cast as language modeling with an implicit prompt (e.g. ``The sentiment in this review is [MASK].''). Conversely, RACE
is a multiple-choice classification task, and indeed its \gof{} is relatively low, despite having a large number of training examples.

An exception to the above is SQuAD 2.0, which presents a worse scaling law compared to  SQuAD 1.1, despite having almost 50\% more training examples. The main difference between the two tasks is their metric and the existence of non-answerable questions in the latter. While SQuAD 1.1 measures simple $F_1$, SQuAD 2.0 evaluates models based on \bestFone. That is, the $F_1$ score reachable if an optimal confidence threshold is chosen \emph{post-hoc} to detect non-answerable questions. 
We conjecture that both the evaluation metric, and the task of detecting non-answerable questions contributed to diverging from the LM objective and thus explain the degradation in the \gof.
To test this, we take the models that were finetuned on SQuAD 2.0 and evaluate them with the SQuAD 1.1 metric on the subset of answerable questions.
This indeed increases $R^2$ from $0.797$ to $0.915$. We provide further details in App.~\ref{app:squad2}.
\looseness=-1

Finally, we note that we do not test whether evaluation loss exhibits a power law on finetuning tasks, since we tune hyperparameters based on the target metric. 
This is since log-loss can increase significantly even when the target metric is still improving, due to a single  example in the evaluation set that incurs higher and higher loss during finetuning \cite{JMLR:v19:18-188}. For further discussion, see App.~\ref{app:loss-fit}.

\paragraph{Critical size}
\label{subsec:critical-size}
A possible reason for low $R^2$ scores is that models need a minimal ``size'' to handle a certain task. We  define $R^2_{2:}$ to be the \gof{} when considering only models with depth of at least $2$. 
As is evident from Table~\ref{table:finetuning-results}, the $R^2$ substantially improves in SQuAD 2.0, RACE and MRPC when omitting the smallest scale.
This finding is in line with recent work \cite{Chowdhery2022PaLMSL,Wei2022EmergentAO,Srivastava2022BeyondTI} which suggests some capabilities of language models may emerge only from a certain scale.

\subsection{Hyperparameter tuning}
\label{subsec:hpt}

\input{table_defs/hpt_effect}

When scaling models, one cannot assume that hyperparameters found for one scale would remain optimal for other scales. Despite this, prior work did not report a hyperparameter tuning (\HPT{}) phase during finetuning.
Table~\ref{table:hpt-effect} highlights the difference in \gof{} between models trained with the hyperparameters used by BERT-Base vs. when tuned for each scale individually. 
As can be seen by the $R^2$ values in the table, when \HPT{} is performed, the power law of scaled models tends to be cleaner. Moreover, because hyperparameters were originally tuned on larger models, smaller scales exhibit a large discrepancy from their optimal performance.
This leads to an imprecise power law fit with a steeper slope compared to when HPT is performed (see Table~\ref{table:hpt-effect}), manifested by overshooting predictions. For details on \HPT{}, see App.~\ref{app:hpt}.

\input{table_defs/predictive_power}

\subsection{Predictive power}
\label{subsec:predictive}

To check whether scaling laws are useful, we need to evaluate their predictive power. To test that, we conduct the following experiments. First, we split the samples used to fit power laws, as discussed above, and test how well do models with 1-6 layers predict the performance of models with 7 or 8 layers (aspect ratio 32), and evaluate with MRE. 

The column MRE in Table~\ref{table:finetuning-results} shows the results of this experiment.
In most tasks, the \emph{MRE} is quite small ($\le2.5\%$), and is correlated with $R^2$. For example, the six smaller models (with 1-6 hidden layers) finetuned on SQuAD 1.1 predict the F$_1$ of the two larger models (7-8 hidden layers) to a 0.6\% relative difference (roughly half an $F_1$ point).
One notable case is SST-2, where MRE is excellent (0.5\%), but $R^2$ is lower than some other tasks (0.951). We attribute this to the fact that the slope of the fitted line in SST-2 is relatively gentle -- all eight scales score in the range 79.4-88.5, see Fig.~\ref{fig:finetuning-results}. Since $R^2$ measures the proportion of variance explained compared to a constant prediction, it is more sensitive to errors when the slope is close to zero.
Similarly, BoolQ also presents a good \emph{MRE} score even though its $R^2$ is low. Analyzing Fig.~\ref{fig:finetuning-results} shows that while the fit is poor, most scores lie in a small range and close to the na\"ive majority baseline, explaining this contradiction.

Expanding on this, we pretrain a 12-layer BERT-Base model from scratch using the same setup as all models, and with MLM as the objective. We then finetune it on the different tasks. Table~\ref{table:finetuning-predictive-power} shows the relative prediction error (\emph{RE}) based on the fit
from models with 1-8 layers. Note that the largest model, with 8 layers, has 14x less parameters compared to BERT-Base. In all cases where the goodness of fit is high ($R^2\ge\RSQTH$), the prediction is accurate to less than 3\% difference and in some cases to almost 0.1\%. Following the discussion in \S\ref{subsec:critical-size} where we saw that RACE, SQuAD 2.0 and MRPC might be affected by a critical size limit, we compute the \emph{RE}$_{2:}$ based on the models with 2-8 layers and report the results in Table~\ref{table:finetuning-predictive-power}. Indeed, all three datasets, which gain a significant boost in $R^2_{2:}$ compared to $R^2$, also benefit improved predictive power.
Interestingly, in 7 out of 9 tasks, our prediction is over-optimistic (i.e. \emph{RE} is negative). 
This hints that our BERT-Base model might be under-trained.

\input{table_defs/pmi_vs_mlm}

\subsection{Case study: MLM vs. PMI objectives}
\label{subsec:model-selection}

As a case study, we simulate the use of scaling laws from the perspective of a resource-constrained researcher introducing a new pretraining objective, such as PMI. Specifically, we imagine
pretraining and finetuning small models with 1-8 layers, for  the PMI and MLM objectives, and then predicting which model will perform better when scaling up to a model with 12 layers, i.e., BERT-Base.

\paragraph{Model selection} Table~\ref{table:pmi-vs-mlm} compares the predicted performance gap vs. the actual performance gap in the BERT-Base results (positive values indicate predicted/actual performance of PMI is higher than MLM and vice versa).
We find that in all cases where $R^2$ is high, the predicted gap holds the same sign as the actual one, suggesting that the predictions are useful for performing model selection.
Moreover, in SQuAD 1.1, QNLI and MNLI, the predicted gap is very accurate. 
Overall, we conclude that when the \gof{}, i.e.,  $R^2$, is high enough, scaling laws present a viable approach for performing model selection without training a large model. We leave for future work to determine if predictions remain accurate when extrapolating over multiple orders-of-magnitude.

\paragraph{Computational efficiency}
We have shown that for \emph{some} NLP tasks, scaling laws can be an effective tool for model selection, and be further used for debugging convergence. However, applying them requires multiple runs across scales for uncertainty estimation and \HPT. Thus, a key question is how much resources are saved with this effort.

To examine this, we perform a theoretical analysis of the FLOPs required to pretrain and finetune the small-scale models vs. the larger ones in our experimental setup, where we extrapolate to a model that is one order-of-magnitude larger.
Following \newcite{Kaplan2020ScalingLF}, we estimate the number of FLOPs for the forward and backward passes with $C\approx 6ND$ where $D$ is the total number of tokens observed. Assuming all models observe the same number of tokens (ignoring early stopping as it requires extra FLOPs for the evaluation set), the difference in computation arises solely from the number of parameters. For example, the total count of parameters of the 8 small-scale models in our setup is 16M while BERT-Base contains 85M, suggesting a 5x improvement. 
Extrapolating to larger scales will yield more substantial savings, but we did not pretrain and evaluate any larger models.

In practice, we observed that smaller scale models require more epochs during finetuning, increasing their token count, $D$.
Still, even if we sum all FLOPs performed for \HPT{} and finetuning over all scales and \emph{all 9 tasks}, we empirically observed a 2.5x decrease in FLOP count compared to training the larger model. We do not compare runtime because different models were trained on different types of nodes, but we expect savings in terms of runtime to be even greater, as training multiple models is trivial to parallelize.

Overall, one can expect compute savings of 2.5-5x when scaling to a model that is an order of magnitude larger, albeit at the cost of performing careful monitoring of convergence and \HPT{}.

%% file: figure_defs/finetuning_results.tex
\begin{figure*}[t]
\begin{center}
\includegraphics[width=\textwidth]{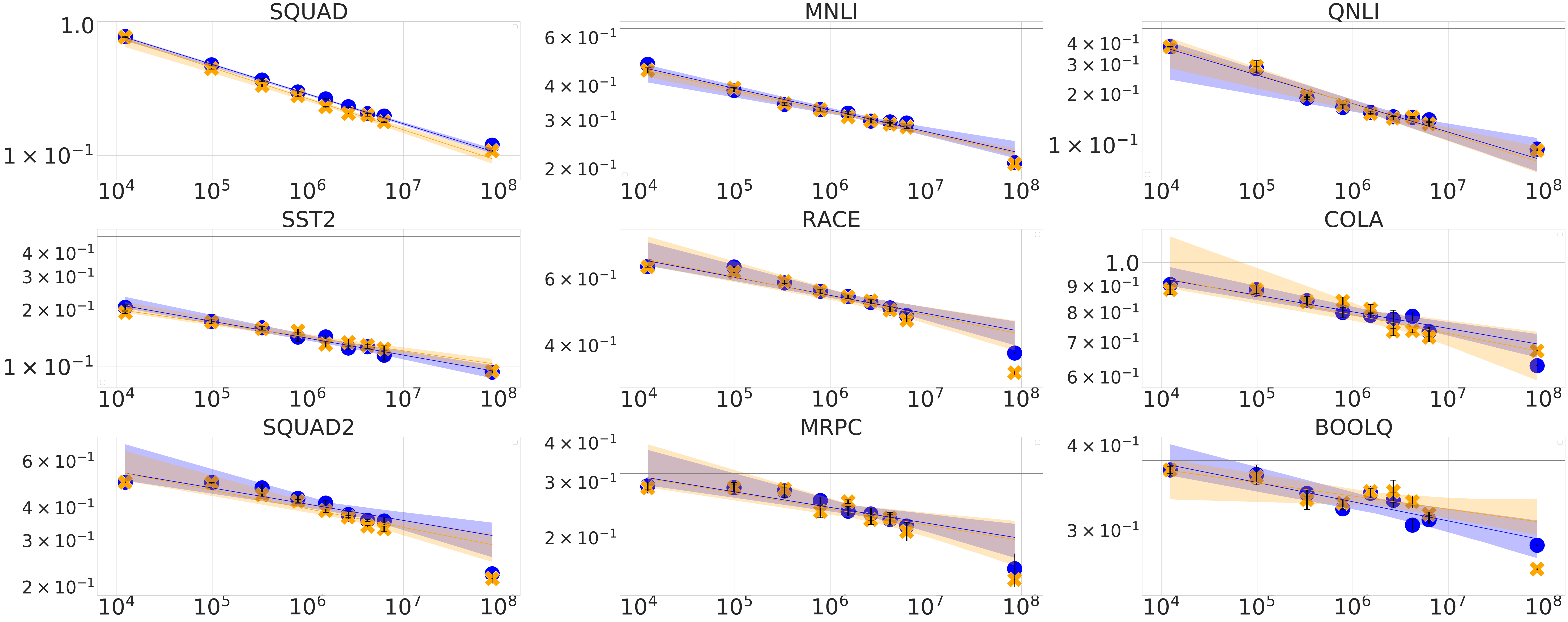}
\setlength{\belowcaptionskip}{0pt}
\setlength{\abovecaptionskip}{-5pt}
\caption{Evaluation performance w.r.t.\ number of parameters. The y-axis of each plot is $1-\mathtt{Metric}$ where $\mathtt{Metric}$ is listed in table~\ref{table:datasets}. The horizontal gray line indicate the simple baseline achieved by a majority vote on the most frequent class, when applicable. \textbf{Blue}: models pretrained with MLM. \textbf{Orange}: models pretrained with PMI.} 
\label{fig:finetuning-results}
\end{center}
\end{figure*}

%% file: table_defs/finetuning_results.tex
\begin{table}[t]
\centering
\footnotesize

\begin{tabular}{@{\extracolsep{10.0pt}}lcc:c} 
\Xhline{2\arrayrulewidth}

  & \boldmath{$R^2$} & \textbf{MRE} & \boldmath{$R^2_{2:}$} \\ \cline{2-2}\cline{3-3}\cline{4-4}
\rule{-2pt}{2.6ex}

\input{figures/finetuning_gof.txt}

\Xhline{2\arrayrulewidth}
\end{tabular}
\setlength{\belowcaptionskip}{0pt}
\caption{
Goodness-of-fit on finetuning tasks. \emph{MRE} measures the predictive power of scales 1-6 ($M=6$), evaluated on scales 7,8. 
The line separates tasks with $R^2\ge\RSQTH$ ($M=8$). $R^2_{2:}$ measures \gof\ on scales 2-8  ($M=7$). In all cases, $T=5$.
}
\label{table:finetuning-results}
\end{table}

%% file: figures/finetuning_gof.txt
 \textbf{SQuAD 1.1} & 0.998 & 0.006 & 0.995 \\
      \textbf{MNLI} & 0.971 & 0.020 & 0.961 \\
      \textbf{QNLI} & 0.965 & 0.024 & 0.876 \\
     \textbf{SST-2} & 0.951 & 0.005 & 0.906 \\
\hdashline
      \textbf{RACE} & 0.916 & 0.045 & 0.974 \\
      \textbf{CoLA} & 0.845 & 0.088 & 0.820 \\
 \textbf{SQuAD 2.0} & 0.797 & 0.057 & 0.917 \\
      \textbf{MRPC} & 0.763 & 0.020 & 0.849 \\
     \textbf{BoolQ} & 0.749 & 0.023 & 0.669 \\

%% file: table_defs/hpt_effect.tex
\begin{table}[t]
\centering
\footnotesize
\addtolength{\tabcolsep}{-2pt}  
\begin{tabular}{@{\extracolsep{-0.0pt}}lcccc} 

\Xhline{2\arrayrulewidth}

 & \multicolumn{2}{c}{\boldmath{$R^2$}} & \multicolumn{2}{c}{\textbf{Slope}} \\\cline{2-3}\cline{4-5}

\rule{-2pt}{2.6ex}

  & \textbf{HPT} & \textbf{w/o HPT} & \textbf{HPT} & \textbf{w/o HPT}     \\ \cline{2-2}\cline{3-3}\cline{4-4}\cline{5-5}
\rule{-2pt}{2.6ex}

\input{figures/hpt_effect.txt}

\Xhline{2\arrayrulewidth}
\end{tabular}

\setlength{\belowcaptionskip}{0pt}
\caption{Effect of performing \hpt\ on the fit. SQD 1 and SQD 2 refer to SQuAD 1.1 and 2.0.}
\label{table:hpt-effect}
\end{table}
\addtolength{\tabcolsep}{2pt}  

%% file: figures/hpt_effect.txt
 \textbf{SQD 1} & 0.998 & 0.830 &  $-0.22 \pm 0.01$ &  $-0.30 \pm 0.07$ \\
  \textbf{MNLI} & 0.971 & 0.933 &  $-0.07 \pm 0.02$ &  $-0.10 \pm 0.03$ \\
  \textbf{QNLI} & 0.965 & 0.867 &  $-0.14 \pm 0.05$ &  $-0.18 \pm 0.08$ \\
 \textbf{SST-2} & 0.951 & 0.960 &  $-0.10 \pm 0.02$ &  $-0.11 \pm 0.02$ \\
  \textbf{RACE} & 0.916 & 0.949 &  $-0.06 \pm 0.02$ &  $-0.06 \pm 0.01$ \\
  \textbf{CoLA} & 0.845 & 0.735 &  $-0.03 \pm 0.01$ &  $-0.05 \pm 0.02$ \\
 \textbf{SQD 2} & 0.797 & 0.765 &  $-0.08 \pm 0.03$ &  $-0.08 \pm 0.04$ \\
  \textbf{MRPC} & 0.763 & 0.740 &  $-0.06 \pm 0.03$ &  $-0.07 \pm 0.03$ \\
 \textbf{BoolQ} & 0.749 & 0.838 &  $-0.03 \pm 0.01$ &  $-0.03 \pm 0.01$ \\

%% file: table_defs/predictive_power.tex
\begin{table}[t]
\centering
\footnotesize

\begin{tabular}{@{\extracolsep{1.0pt}}lccc:c} 

\Xhline{2\arrayrulewidth}

  & \textbf{Prediction} & \textbf{Actual} & \textbf{RE} & \textbf{RE}$_{2:}$ \\ \cline{2-2}\cline{3-3}\cline{4-4}\cline{5-5}

\rule{-2pt}{2.6ex}
\input{figures/finetuning_predictive_power.txt}

\Xhline{2\arrayrulewidth}
\end{tabular}
\setlength{\belowcaptionskip}{0pt}
\setlength{\belowcaptionskip}{-5pt}
\caption{
Predictive power based on 8 smaller models, predicting the performance of BERT-Base. The dashed line separates tasks with $R^2\ge\RSQTH$, and \emph{RE}$_{2:}$ is the \emph{RE} when the fit ignores single layer models.
}
\label{table:finetuning-predictive-power}
\end{table}

%% file: figures/finetuning_predictive_power.txt
 \textbf{SQD 1} & 0.893 & 0.880 &  0.014 &  0.011 \\
  \textbf{MNLI} & 0.769 & 0.790 & -0.026 & -0.040 \\
  \textbf{QNLI} & 0.917 & 0.905 &  0.013 &  0.008 \\
 \textbf{SST-2} & 0.905 & 0.906 & -0.002 &  0.001 \\
\hdashline
  \textbf{RACE} & 0.561 & 0.618 & -0.092 & -0.045 \\
  \textbf{CoLA} & 0.305 & 0.369 & -0.175 & -0.125 \\
 \textbf{SQD 2} & 0.685 & 0.776 & -0.117 & -0.076 \\
  \textbf{MRPC} & 0.800 & 0.841 & -0.049 & -0.030 \\
 \textbf{BoolQ} & 0.708 & 0.715 & -0.009 &  0.000 \\

%% file: table_defs/pmi_vs_mlm.tex
\begin{table}[t]
\centering
\footnotesize
\addtolength{\tabcolsep}{-2pt}  
\begin{tabular}{@{\extracolsep{-0.0pt}}lcccccc} 

\Xhline{2\arrayrulewidth}

  & \textbf{Predicted}\ \boldmath{$\Delta$} & \textbf{Actual}\ \boldmath{$\Delta$} & \textbf{Sign($\Delta$)}    &
  \boldmath{$R^2$} \\
  \cline{2-2}\cline{3-3}\cline{4-4}\cline{5-5}

\rule{-2pt}{2.6ex}

\input{figures/pmi_vs_mlm.txt}

\Xhline{2\arrayrulewidth}
\end{tabular}
\setlength{\belowcaptionskip}{0pt}
\caption{Finetuning results for models pretrained with MLM/PMI. Positive $\Delta$ indicates PMI is better than MLM. Tasks above line have $R^2\ge\RSQTH$.}
\label{table:pmi-vs-mlm}
\end{table}
\addtolength{\tabcolsep}{2pt}

%% file: figures/pmi_vs_mlm.txt
 \textbf{SQD 1} &  0.013 &  0.011 &   Correct  & $\ge\RSQTH$\\
  \textbf{MNLI} &  0.002 &  0.002 &   Correct & $\ge\RSQTH$\\
  \textbf{QNLI} &  0.003 &  0.002 &   Correct & $\ge\RSQTH$\\
 \textbf{SST-2} & -0.009 & -0.001 &   Correct & $\ge\RSQTH$\\
\hdashline
  \textbf{RACE} &  0.007 &  0.043 &   Correct & $<\RSQTH$\\
  \textbf{CoLA} &  0.021 & -0.043 &   Wrong & $<\RSQTH$\\
 \textbf{SQD 2} &  0.024 &  0.009 &   Correct & $<\RSQTH$\\
  \textbf{MRPC} &  0.003 &  0.012 &   Correct & $<\RSQTH$\\
 \textbf{BoolQ} & -0.017 &  0.022 &   Wrong & $<\RSQTH$\\

%% file: 05_related_work.tex
\section{Related Work}

\paragraph{Scaling laws in transformers} 
\label{subsec:scaling} Since \newcite{Kaplan2020ScalingLF} demonstrated scaling laws for transformer-based language models, researchers have been investigating the extent of this phenomenon. 
\newcite{Tay2022ScalingLV} investigated the effect of inductive bias on scaling laws and showed how different architectures affect the emerging scaling laws. \newcite{Tay2021ScaleEI} showed that model shape matters, and that pretraining and finetuning losses are not necessarily correlated. Contrary to \newcite{Kaplan2020ScalingLF}, they showed that shape also plays a role in finetuning performance, rather than size alone. \newcite{Hernandez2021ScalingLF} focused on python code, showing a trade-off between data and compute. \newcite{Ghorbani2021ScalingLF} analyzed scaling laws in transformer models used in neural machine translation while \newcite{gordon-etal-2021-data} and \newcite{Bansal2022DataSL} discussed practical implications of the predictive power of such results. 

In parallel to work done in NLP, several works focused on scaling laws in other domains. \newcite{Henighan2020ScalingLF} observed scaling laws in multi-modal settings. \newcite{Zhai2021ScalingVT} gave a comprehensive review on scaling laws behavior of upstream computer vision tasks, and \newcite{abnar2021exploring} investigated the relationships between upstream and downstream performance of vision transformers.

An important line of work was dedicated to explaining the emergence of scaling laws in neural models \cite{Hashimoto2021ModelPS}. \newcite{Bahri2021ExplainingNS} connect the scaling exponent to the intrinsic dimension of the data-manifold realized by trained networks representations. \newcite{Bordelon2020SpectrumDL} and \newcite{Bisla2021ATA} connect scaling behavior to the spectrum of the kernel in the related NTK model. Theoretical explanations for neural scaling laws include analogy to kernel methods \cite{Spigler2020AsymptoticLC,Bordelon2020SpectrumDL}, nearest neighbors methods \cite{Sharma2022ScalingLD,Bisla2021ATA}, or a combination of the two \cite{Bahri2021ExplainingNS}.

\paragraph{Harnessing scaling laws for model design} \newcite{Rosenfeld2020ACP} performed small-scale experiments to approximate the generalization error of larger models with a functional form accounting for model and dataset sizes. However, they focused on pretraining, while we also investigate the predictive power on downstream language tasks. 

\newcite{Hashimoto2021ModelPS} used scaling laws to predict the optimal composition of a training set from different data sources. \newcite{Kirstain2021AFM} investigated the effect of parameter count and data size on improving performance on various language tasks, while \newcite{johnson-etal-2018-predicting} designed a performance extrapolation task to estimate how much training data is needed to achieve the required performance.

A parallel line of work that tries to extrapolate optimal architectures based on small scale experiments is \emph{Neural Architecture Search} \cite{Zoph2017NeuralAS}. Such methods have outperformed human designed architectures \cite{Zoph2017NeuralAS,Liu2019DARTSDA,Chen2018SearchingFE,Real2019RegularizedEF}. However, it has been shown that the resulting architectures, such as EfficietNet \cite{Tan2019EfficientNetRM}, do not always scale well \cite{Bello2021RevisitingRI}.

%% file: 06_discussion.tex
\section{Discussion and Future Work}

In this work, we show that scaling laws can be used as an effective tool for model selection and as a diagnostic tool to test convergence of large-scale models. Our \emph{practical takeaways} are:
\begin{enumerate}%[leftmargin=*,itemsep=0pt,topsep=0pt]
\item 
Scaling laws can be used as an effective predictive and diagnostic tool, as long as the fit is good. Specifically, we find that $R^2\ge\RSQTH$ is a good indicator.
\item Performing independent \HPT{} at every scale is crucial for model selection and for the emergence of scaling laws in particular.
\item Pretraining models to convergence is important for observing the scaling behavior of transformer models over downstream NLP tasks.
\end{enumerate}

%% file: 07_conclusions.tex
\section{Conclusion}
This work is motivated by a practical question: 
can scaling laws provide a principled method for developing models at very small scale and extrapolating to larger ones?
We perform a thorough empirical analysis on the emergence of scaling laws on a wide range of language understanding tasks. 
We find that scaling laws emerge for \emph{some} tasks, potentially as a function of the proximity between the downstream task and the pretraining objective, but revealing them incurs the overhead of hyperparameter tuning across multiple scales. Our results show that scaling laws are beneficial for debugging model convergence when training large models, and to predict model behavior when they emerge at small scale. \looseness=-1

%% file: 08_limitations.tex
\section{Limitations}
We discuss four limitations left for future work. (a) we focused on small-scale models, and thus have no empirical evidence for models that are significantly larger; (b) we analyze encoder-only modes, and leave decoder-based models for future work; (c)
we analyze 9 different downstream tasks, they are all based on English-only datasets, and none of them evaluate models' generation capabilities.
(d) while we provide a rule-of-thumb for telling if scaling laws predictions are reliable, it remains unclear why scaling laws do not apply to all downstream tasks, which remains an area for future research.

%% file: 09_acknowledgements.tex
\section*{Acknowledgements}
We thank Mor Geva for her useful comments. This research was partially supported by The Yandex Initiative for Machine Learning, the Shashua Fellowship, 
the Len Blavatnik and the Blavatnik Family foundation, the 
Israeli Science Foundation (ISF) grant no. 2486/21,
and the European Research Council (ERC) under the European Union Horizons 2020 research and innovation programme (grant ERC DELPHI 802800). This work was completed in partial fulfillment for the Ph.D. degree of the first author.

%% file: appendix_00_main.tex
\clearpage  % https://tex.stackexchange.com/a/2960
\setcounter{page}{0}
\pagenumbering{arabic}
\setcounter{page}{1}

\section{Appendix}

\input{appendix_01_experimental}

\input{appendix_02_hpt}
\input{appendix_03_evalaution}

\input{appendix_04_convergence}
\input{appendix_05_mlm_pmi}
\input{appendix_06_loss_fit}
\input{appendix_07_squad2}

%% file: appendix_01_experimental.tex
\subsection{Experimental details}
\label{app:experimental-details}

All experiments were done with the \emph{transformers} \cite{wolf-etal-2020-transformers} library (version 4.4.0) and tracked using the \emph{Comet.ML} infrastructure \cite{CometML}. All pretraining and finetuning datasets were provided by the \emph{Datasets library} \cite{lhoest-etal-2021-datasets} (version 1.4.1) and were left as is, except for MNLI \cite{Williams2018ABC} where a random subset of 25\% of the training samples were used to finetune the models. For pretraining, we used the Wikipedia dump provided by \emph{dataset} library in the English dataset configuration name \textit{20200501.en}. Whenever possible, example recipes provided by \emph{transformers} where used to tune the models. We trained all models until convergence (discussed further in App.~\ref{app:convergence}) and chose the checkpoint that performed best over the evaluation set (i.e., post-hoc early stopping). To support the training and analysis of the results, we used \emph{numpy} \cite{harris2020array}, \emph{scipy} \cite{2020SciPy-NMeth}, \emph{pandas} \cite{mckinney-proc-scipy-2010,reback2020pandas} and \emph{scikit-learn} \cite{scikit-learn}. All models ran using \emph{PyTorch} \cite{NEURIPS2019_9015}.

The complete configuration of the different scales can be found in Table~\ref{table:model-configurations}. In all cases, we focused on the emergence of scaling laws in NLP tasks rather than achieving optimal results, and thus did not perform any ``mid-training'' \cite{Wang2018GLUEAM}.
To account for randomness, during finetuning, we used five different seeds for each model-task pair. However, since pretraining is more expensive, we only have a single random seed during pretraining.

\input{table_defs/model_configurations}

%% file: table_defs/model_configurations.tex
\begin{table}[ht]
\centering

\begin{tabular}{@{\extracolsep{3.0pt}}cccc} 

\Xhline{2\arrayrulewidth}

  \textbf{AR} & \textbf{Layers} & \textbf{Heads} & \boldmath{$N$} \\ \cline{1-1} \cline{2-2}\cline{3-3}\cline{4-4}\cline{1-1}
\rule{-2pt}{2.6ex}

\input{figures/models_configuration.txt}

\Xhline{2\arrayrulewidth}
\end{tabular}
\caption{Model configurations and number of parameters. The last line represents BERT-Base \cite{Devlin2019BERTPO}. 
}
\label{table:model-configurations}
\end{table}

%% file: figures/models_configuration.txt
32 &  1 &  4 &     12,288 \\
32 &  2 &  4 &     98,304 \\
32 &  3 &  4 &    331,776 \\
32 &  4 &  4 &    786,432 \\
32 &  5 &  4 &  1,536,000 \\
32 &  6 &  4 &  2,654,208 \\
32 &  7 &  4 &  4,214,784 \\
32 &  8 &  4 &  6,291,456 \\
64 &  1 &  4 &     49,152 \\
64 &  2 &  4 &    393,216 \\
64 &  3 &  4 &  1,327,104 \\
64 &  4 &  4 &  3,145,728 \\
64 &  5 &  4 &  6,144,000 \\
64 & 12 & 12 & 84,934,656 \\

%% file: appendix_02_hpt.tex
\input{figure_defs/ft_hpt_effect}

\subsection{Hyperparameter tuning}
\label{app:hpt}
\paragraph{Finetuning}
As discussed in \S\ref{subsec:hpt}, we observed that \hpt{} during finetuning has considerable impact. To determine the hyperparameters to tune, we first experimented with modifying the different options in various scales and tasks. In particular, we examined the effects of \emph{weight decay}, \emph{batch size}, \emph{number of epochs}, \emph{initial learning rate}, \emph{warm-up}, \emph{learning rate scheduler} and \emph{dropout}. We found that by fixing the \emph{learning rate scheduler} to \emph{CONSTANT}, the dropout rate and warm-up to 10\% and the \emph{batch size} to high enough (64), we are able to outperform other configurations by only varying the \emph{learning rate} and \emph{total number of epochs}. Then, we started performing a grid-search to choose those values for each model-task pair. Figure.~\ref{fig:ft-hpo-effect} shows the effect of this \hpt.

\paragraph{Pretraining}
In the case of pretraining, we have found that by using a large enough \emph{batch size} (256) and a small enough initial \emph{learning rate} ($10^{-4}$) with long training horizon (500K steps) all models achieve comparable results at convergence to those achieved when hyperparameters were tuned (though the convergence rate differed). The rest of the hyperparameters used were a \emph{linear learning rate scheduler}, 0.1 \emph{dropout rate} and 10K \emph{warmup} steps.

%% file: figure_defs/ft_hpt_effect.tex
\begin{figure*}[t]
\begin{center}
\centerline{
\includegraphics[width=\linewidth]{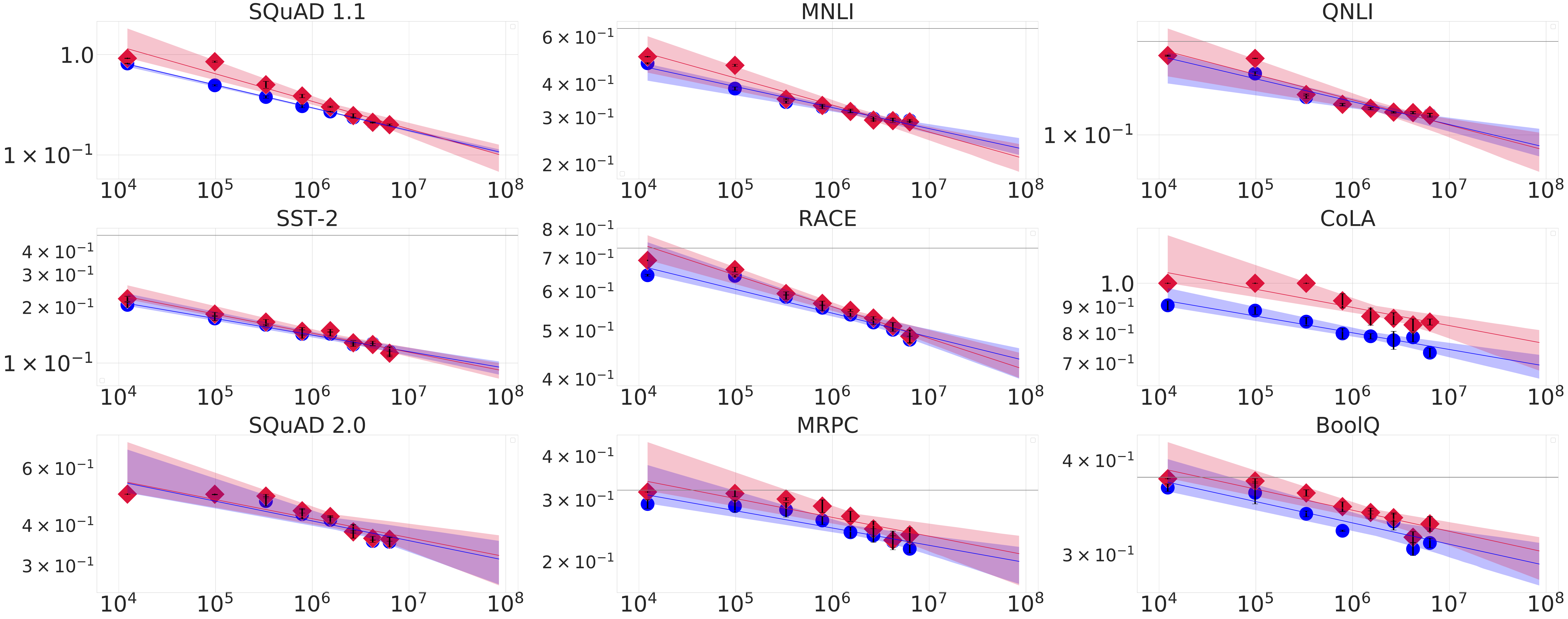}
}
\setlength{\belowcaptionskip}{-5pt}
\caption{Effect of \HPT{} on finetuning scaling laws (axes as in figure~\ref{fig:finetuning-results}). \textbf{Blue}: with \HPT{}. \textbf{Red}: w/o \HPT{}.
% \mi{squuezed it}
} 
\label{fig:ft-hpo-effect}
\end{center}
\end{figure*}

%% file: appendix_03_evalaution.tex
\input{figure_defs/figure_bootstrapping}
\input{algorithm_defs/bootstrap_algorithm}
\subsection{Evaluation}
\label{app:evaluation}

As discussed and is visible in Fig.\ref{fig:finetuning-results}, there is variance in the finetuning performance resulting from the different seeds. Thus, we fit the power laws based on multiple random seeds to capture the \textbf{expected} performance of a scale. Moreover, 
we compute the uncertainty of the fit in the form of a confidence interval. To do so, we use our data points to bootstrap $B=1000$ fitted lines, and take the $[2.5,97.5]$ percentiles. However, we find that the na\"ive bootstrapping approach in which one samples $b$ points with replacements from a pool of size $b$ (where $b:=M\times T$) only accounts for the variance coming from the finetuning seeds. Since we expect more variance to come from different pretraining random seeds (as was shown by \newcite{Zhong2021AreLP}), we use a hierarchical bootstrapping algorithm (Alg.~\ref{alg:bootstrap}) to compensates for this. 
Our hierarchical bootstrapping procedure works by first sampling $M$ scales with replacement, and then for each scale, we sample $T$ data points with replacement. While the number of observations in each fitted sample is the same, most samples will not include all scales and instead give more weight to specific scales when fitting a power law. When referring to the subroutine $\mathtt{FitLine}(p_1,\dots,p_k)$ where each point $p_i=(x_i,y_i)$, we fit a power law function such that $ln(y)=\alpha\cdot ln(x) +\beta$ and minimize the squared loss $\sum_{i}{(ln(y_i)-ln(\hat{y_i}))^2}$. When computing $R^2$, we use the same procedure over all $M\times T$ points to fit a line, and compute the \gof\ w.r.t.\ to its predictions.
Specifically, given a fitted line $f: \mathbb{R}^+\rightarrow\mathbb{R}^+$, $R^2$ is defined as:
\begin{equation}
    R^2 := 1 - \frac{\mathtt{SS_{res}}}{\mathtt{SS_{tot}}}
\end{equation}
where
\begin{equation}
    \mathtt{SS_{res}} := \sum_{i}{(y_i-f(x_i))^2}
\end{equation}
\begin{equation}
    \mathtt{SS_{tot}} := \sum_{i}{(y_i-\bar{y})^2}
\end{equation}
such that $\bar{y}=\frac{1}{k}\sum_{i=1}^{k}{y_i}$.

As can be seen in Fig.~\ref{fig:bootsrapping}, our approach results in considerably more conservative confidence intervals, where the only difference comes from pretraining variance. We further use this estimation to get the range (in the same percentiles) of the slope, and thus can measure how uncertain the prediction is.

%% file: figure_defs/figure_bootstrapping.tex
\begin{figure*}[t]
\begin{center}
\includegraphics[width=\textwidth]{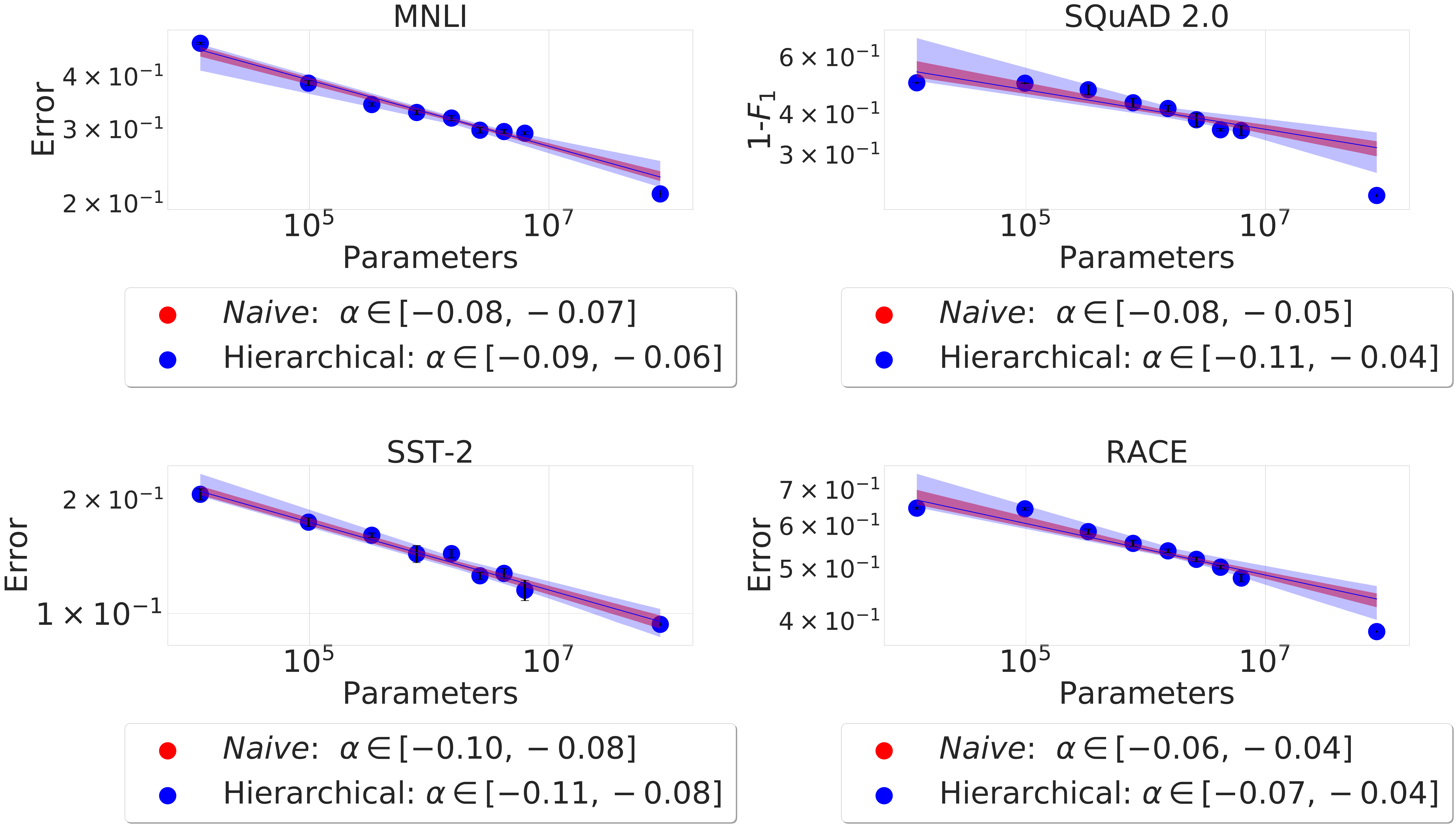}
\caption{Comparison of our hierarchical bootstrapping method against the na\"ive (i.e. ``flat'') sampling approach.} 
\label{fig:bootsrapping}
\end{center}
\end{figure*}

%% file: algorithm_defs/bootstrap_algorithm.tex
\begin{algorithm}[t]
\footnotesize
\SetAlgoLined
    \SetKwInOut{Input}{input}
    \SetKwInOut{Output}{output}
    \Input{performance results $p_{i,j}$ where $i \in [1,M]$ and $j \in [1,T]$, 
    number of samples $B$}
    \Output{$(\alpha_1,\beta_1),..,(\alpha_B,\beta_B)$ \# $b$ fitted lines} 
    \For{$b=1, \dots ,B$}
    {
        \# Sampling uniformly with replacement \\
        $\mathtt{scales} \leftarrow$ sample $M$ scales from $[1,M]$ \\
        $\mathtt{points} \leftarrow [\ ]$ \\
        \For{$i \in \mathtt{scales}$} {
            $\mathtt{sp} \leftarrow$ sample $T$ points from
            ${p_{i,1},...,p_{i,T}}$ \\
            $\mathtt{points.extend}(sp)$
        }
        $(\alpha_i,\beta_i) \leftarrow \mathtt{FitLine(points)}$
    }
    \Return $(\alpha_1,\beta_1), \dots,(\alpha_B,\beta_B)$
 \caption{Hierarchical bootstrap}
 \label{alg:bootstrap}
\end{algorithm}

%% file: appendix_04_convergence.tex
\input{figure_defs/figure_early_stopping}
\subsection{Model convergence}
\label{app:convergence}
We discussed the importance of controlling for variance and tuning hyperparameters when evaluating scaling laws. However, another potential pitfall is model under-training. While many past lines of work report that they ``train until convergence'', they do not explicitly discuss the criteria for stopping training. In particular, when using a decaying learning rate schedule (e.g., \emph{Linear}), the change in loss will tend to 0 as the learning rate approaches zero, which will give the appearance of convergence.
Moreover, when using early-stopping, hyperparameters such as \emph{patience} and \emph{minimal decrease} may affect the final model and lead to under-optimization.

To optimize results, during finetuning we trained for the entire allocated epoch budget and chose post-hoc the best performing checkpoint w.r.t.\ the evaluation metric. However, as pretraining is considerably more compute-intensive, we used ``early stopping'', requiring no decrease in evaluation loss over 1500 consecutive update steps. The only exception is BERT-Base \cite{Devlin2019BERTPO}, where we increased the patience from 1500 to 7500 after suspecting the model may be under-trained.
As can be seen in Fig.~\ref{fig:bert-early-stopping} which shows the evaluation loss of our BERT-Base  over the pretraining data, there are many cases in which the model stops improving for a considerable amount of time (and indeed stops if the early stopping patience is set too low) despite not converging yet. 
%This is because the method is highly local. 
When "zooming-out", it is clear that the model is still training, and has not reached convergence. This supports our hypothesis that our large scale model is under-trained and accounts for some of the error in the predictions from Table~\ref{table:finetuning-results} and Fig.~\ref{fig:finetuning-results}. 

%% file: figure_defs/figure_early_stopping.tex
\begin{figure}[ht]
\begin{center}
\includegraphics[width=\linewidth]{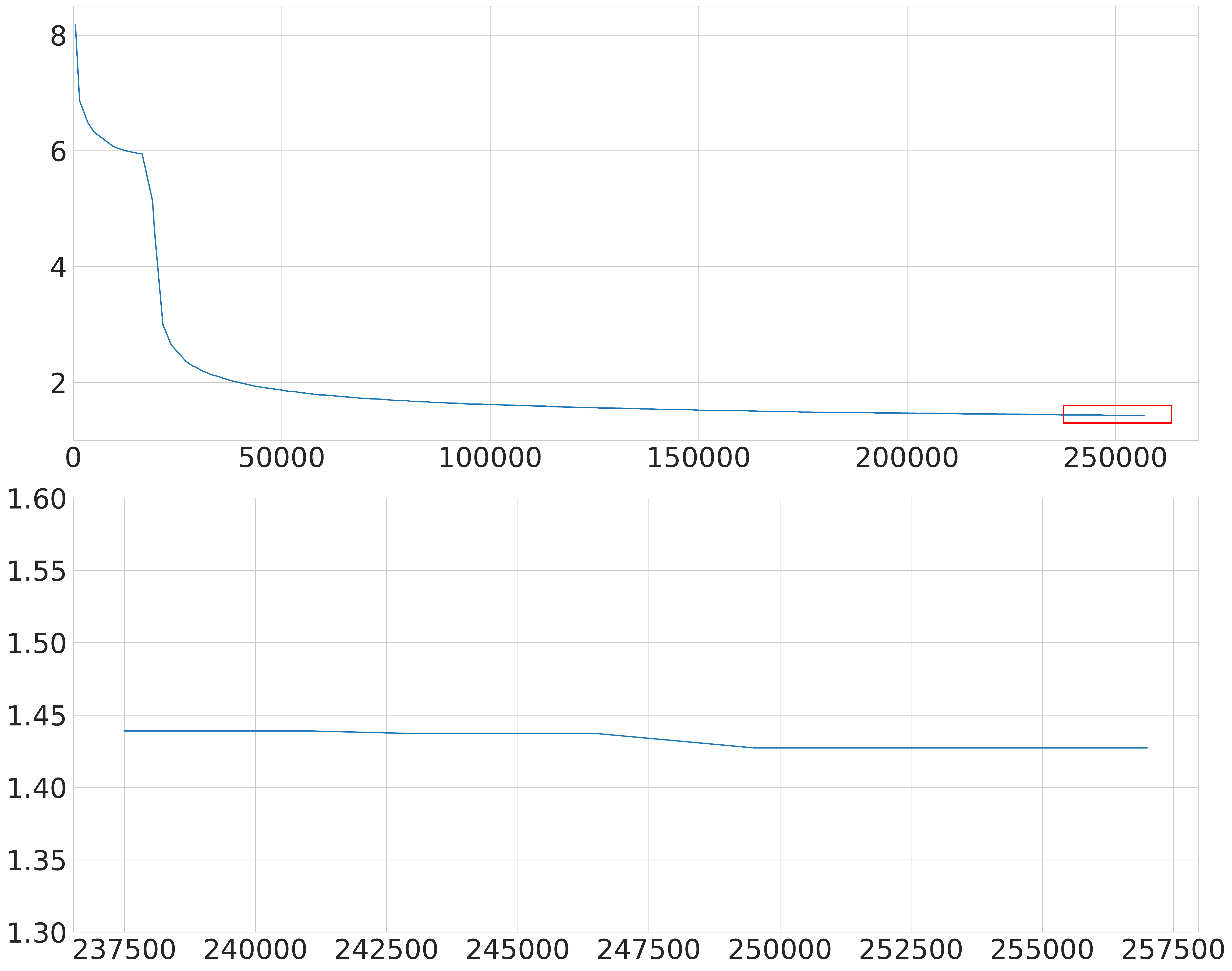}
\caption{Evaluation loss of BERT-Base \cite{Devlin2019BERTPO} pretraining with MLM against the number of training steps. \textbf{Bottom}: Zoom-in on the final steps.
} 
\label{fig:bert-early-stopping}
\end{center}
\end{figure}

%% file: appendix_05_mlm_pmi.tex
\input{table_defs/pmi_mlm_appendix}

\subsection{PMI vs. MLM results}
\label{app:performance}
Table~\ref{table:pmi-mlm-r2} shows the full set of scaling law results for the various models and tasks, comparing pretraining with MLM and PMI. 
While \S\ref{subsec:model-selection} discusses the potential benefits of using scaling laws as a method for model selection, we observe a slight disparity between our BERT-Base scores and the ones reported in literature. In particular, \newcite{levine2020pmi} reported their BERT-Base models trained with PMI to achieve $81.4$ \bestFone\ score on SQuAD 2.0 and $70.1$ accuracy on RACE when pretraining for $1M$ steps. We attribute the difference to the significantly more update steps taken ($1M$ vs. our $250K$) and usage of the \emph{Book Corpus} dataset \cite{Zhu2015AligningBA} during pretraining. This conjecture is supported by the scores they present when pretraining an additional $1.4M$ steps ($83.3$ and $72.3$ respectively) as well as increasing the pretraining corpora significantly ($83.9$ and $74.8$ respectively).   

%% file: table_defs/pmi_mlm_appendix.tex
\begin{table*}[t]
\centering
\footnotesize
\addtolength{\tabcolsep}{-2pt}  
\resizebox{\linewidth}{!}{\begin{tabular}{@{\extracolsep{-0.0pt}}lcccccc:cccc} 

\Xhline{2\arrayrulewidth}

 & \multicolumn{2}{c}{\boldmath{$R^2$}} & \multicolumn{2}{c}{\textbf{Slope}} & \multicolumn{2}{c}{\textbf{Pred. (act.)}} & \multicolumn{2}{c}{\boldmath{$R^2_{2:}$}}
 & \multicolumn{2}{c}{\textbf{RE}$_{2:}$}
 \\\cline{2-3}\cline{4-5}\cline{6-7}\cline{8-9}\cline{10-11}

\rule{-2pt}{2.6ex}

  & \textbf{MLM} & \textbf{PMI} & \textbf{MLM} & \textbf{PMI} & \textbf{MLM} & \textbf{PMI} & \textbf{MLM} & \textbf{PMI} & \textbf{MLM} & \textbf{PMI}      \\ \cline{2-2}\cline{3-3}\cline{4-4}\cline{5-5}\cline{6-6}\cline{7-7}\cline{8-8}\cline{9-9}\cline{10-10}\cline{11-11}
\rule{-2pt}{2.6ex}

\input{figures/pmi_mlm_appendix.txt}

\Xhline{2\arrayrulewidth}
\end{tabular}}
\caption{Comparison of finetuning results for models pretrained with MLM/PMI. $R^2$ is the \gof\ observed on the models with 1-8 layers and the \emph{Slope} is the 95\% confidence interval estimated for the slope of the fit. \emph{Pred.} refer to the prediction of the BERT-Base perforamnce based on the fit, and \emph{act.} is the actual value observed. $R^2_{2:}$ and \emph{RE}$_{2:}$ is the \gof\ and \emph{RE} (see \S\ref{subsec:eval}) respectively, when fitting based only on models with 2-8 layers. SQD 1 and 2 refer to SQuAD 1.1 and 2.0 respectively.}
\label{table:pmi-mlm-r2}
\end{table*}
\addtolength{\tabcolsep}{2pt}

%% file: figures/pmi_mlm_appendix.txt
 \textbf{SQD 1} & 1.00 & 0.99 &  [-0.23, -0.21] &  [-0.25, -0.21] &  0.89 (0.88) &  0.91 (0.89) & 1.00 & 0.99 &  0.01 &  0.02 \\
  \textbf{MNLI} & 0.97 & 0.99 &  [-0.09, -0.06] &  [-0.08, -0.07] &  0.77 (0.79) &  0.77 (0.79) & 0.96 & 0.98 & -0.04 & -0.02 \\
  \textbf{QNLI} & 0.96 & 0.96 &  [-0.20, -0.09] &  [-0.20, -0.12] &  0.92 (0.91) &  0.92 (0.91) & 0.88 & 0.89 &  0.01 &  0.02 \\
 \textbf{SST-2} & 0.95 & 0.93 &  [-0.11, -0.08] &  [-0.09, -0.07] &  0.90 (0.91) &  0.90 (0.91) & 0.91 & 0.89 &  0.00 & -0.01 \\
  \textbf{RACE} & 0.92 & 0.90 &  [-0.07, -0.04] &  [-0.08, -0.04] &  0.56 (0.62) &  0.57 (0.66) & 0.97 & 0.96 & -0.05 & -0.04 \\
  \textbf{CoLA} & 0.85 & 0.75 &  [-0.05, -0.02] &  [-0.07, -0.02] &  0.30 (0.37) &  0.33 (0.33) & 0.82 & 0.84 & -0.12 & -0.01 \\
 \textbf{SQD 2} & 0.80 & 0.86 &  [-0.11, -0.04] &  [-0.11, -0.05] &  0.69 (0.78) &  0.71 (0.79) & 0.92 & 0.98 & -0.08 & -0.04 \\
  \textbf{MRPC} & 0.76 & 0.68 &  [-0.09, -0.03] &  [-0.10, -0.03] &  0.80 (0.84) &  0.80 (0.85) & 0.85 & 0.77 & -0.03 & -0.02 \\
 \textbf{BoolQ} & 0.75 & 0.60 &  [-0.04, -0.02] &  [-0.03, -0.00] &  0.71 (0.71) &  0.69 (0.74) & 0.67 & 0.37 &  0.00 & -0.03 \\

%% file: appendix_06_loss_fit.tex
\subsection{Fitting Evaluation Loss}
\label{app:loss-fit}

\input{figure_defs/finetuning_results_loss}
As mentioned in \S\ref{subsec:born-equal}, the \gof{} for evaluation loss can be quite poor, especially for classification tasks. This is since a single outlier that the model is wrong on with high confidence (and confidence tends to increase during training) can lead to high evaluation log-loss even when the target metric keeps improving; this is a known phenomenon \cite{JMLR:v19:18-188}. As is visible from Fig.~\ref{fig:finetuning-results-loss}, fitting a power-law on the best evaluation loss have no advantage over using the target metric. Specifically, when fitting the evaluation loss rather than evaluation accuracy causes an average drop of over 10\% in \gof{} as measured by $R^2$.

%% file: figure_defs/finetuning_results_loss.tex
\begin{figure*}[t]
\begin{center}
\includegraphics[width=\textwidth]{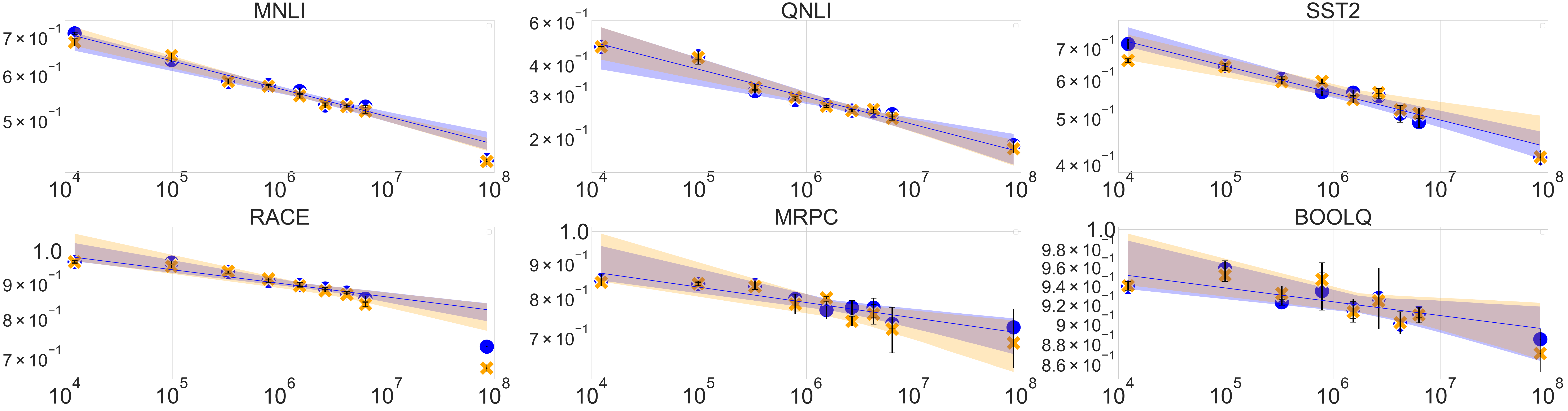}
\setlength{\belowcaptionskip}{0pt}
\setlength{\abovecaptionskip}{-5pt}
\caption{Evaluation loss w.r.t.\ number of parameters. The y-axis of each plot is $\frac{\mathcal{L}_{\mathtt{eval}}}{-ln(c)}$ where $\mathcal{L}_{\mathtt{eval}}$ if the best evaluation loss and $c$ is the number of classes in each $c$-way classification task. \textbf{Blue}: models pretrained with MLM. \textbf{Orange}: models pretrained with PMI.} 
\label{fig:finetuning-results-loss}
\end{center}
\end{figure*}

%% file: appendix_07_squad2.tex
\subsection{SQuAD 2.0}
\label{app:squad2}

\input{figure_defs/squad_metrics}
In \S\ref{subsec:born-equal} we discussed the sub-optimal fit exhibited by SQuAD 2.0 compared to SQuAD 1.1. To test our hypothesis that this is due to the change in metric, we evaluated the models that were trained on SQuAD 2.0 using the SQuAD 1.1 objective.
As expected, Fig.~\ref{fig:squad-metrics} shows they exhibit a much cleaner scaling law (compare red to orange line), with a slope similar to the one observed for SQuAD 1.1 (blue line). The slight difference in the intercept and uncertainty may be attributed to the difference in the dataset itself and the finetuning objectives. As mentioned, the \gof{} measured by $R^2$ increased significantly as well, supporting our hypothesis.

%% file: figure_defs/squad_metrics.tex
\begin{figure}[t]
\begin{center}
\centerline{
\includegraphics[width=\linewidth]{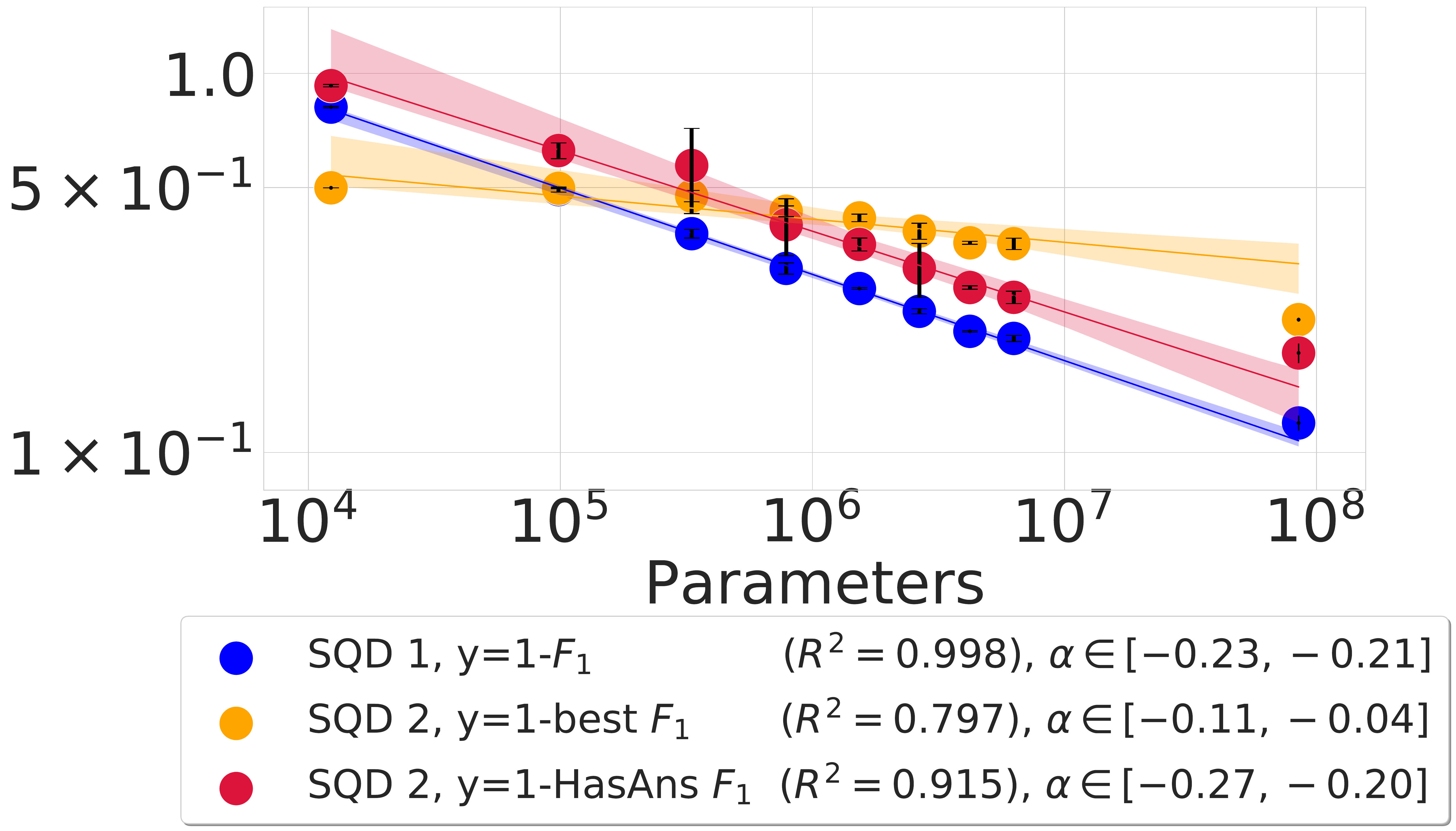}
}
\setlength{\belowcaptionskip}{-5pt}
\caption{Effect of evaluation metric on scaling laws in SQuAD 1.1 and 2.0 (SQD 1 and SQD 2 respectively). SQuAD 2.0 models had their hyperparametrs tuned to minimize $1-$\bestFone, and evaluated also with $F_1$ on the subset of answerable questions from SQuAD 1.1. ($1-$ HasAns $F_1$)} 
\label{fig:squad-metrics}
\end{center}
\end{figure}